\documentclass{article}

\usepackage[preprint,nonatbib]{neurips_2023}

\usepackage[utf8]{inputenc} % allow utf-8 input
\usepackage[T1]{fontenc}    % use 8-bit T1 fonts
\usepackage{hyperref}       % hyperlinks
\usepackage{url}            % simple URL typesetting
\usepackage{booktabs}       % professional-quality tables
\usepackage{amsfonts}       % blackboard math symbols
\usepackage{nicefrac}       % compact symbols for 1/2, etc.
\usepackage{microtype}      % microtypography
\usepackage[table]{xcolor}         % colors
\usepackage{bm}
\usepackage{amsmath}
\usepackage{hyperref}
\usepackage{multirow}
\usepackage{multicol}
\usepackage{graphicx}
\usepackage{algorithm}
\usepackage{algorithmic}
\usepackage{subcaption}
\usepackage{booktabs}
\usepackage{pifont}
\usepackage[square,numbers]{natbib}

\definecolor{lightlightgray}{gray}{0.95}

\newcommand{\cmark}{\ding{51}}%
\newcommand{\xmark}{\ding{55}}%

\title{Continual Learning of Numerous Tasks from Long-tail Distributions}

\author{Liwei~Kang \quad Wee~Sun~Lee \\
National University of Singapore\\
\texttt{\{kang,~leews\}@comp.nus.edu.sg}}

\begin{document}

\maketitle

\begin{abstract}
Continual learning, an important aspect of artificial intelligence and machine learning research, focuses on developing models that learn and adapt to new tasks while retaining previously acquired knowledge. Existing continual learning algorithms usually involve a small number of tasks with uniform sizes and may not accurately represent real-world learning scenarios. In this paper, we investigate the performance of continual learning algorithms with a large number of tasks drawn from a task distribution that is long-tail in terms of task sizes. We design one synthetic dataset and two real-world continual learning datasets to evaluate the performance of existing algorithms in such a setting. Moreover, we study an overlooked factor in continual learning, the optimizer states, e.g. first and second moments in the Adam optimizer, and investigate how it can be used to improve continual learning performance. We propose a method that reuses the optimizer states in Adam by maintaining a weighted average of the second moments from previous tasks. We demonstrate that our method, compatible with most existing continual learning algorithms, effectively reduces forgetting with only a small amount of additional computational or memory costs, and provides further improvements on existing continual learning algorithms, particularly in a long-tail task sequence.

\end{abstract}

\section{Introduction}
Continual learning, also known as lifelong learning, is a critical aspect of artificial intelligence (AI) and machine learning (ML) research which aims to develop models that can learn and adapt to new tasks or information while retaining previously acquired knowledge \citep{mccloskey1989catastrophic,french1999catastrophic}. This capability is essential for creating intelligent agents that can operate in dynamic environments and mimic human-like learning. However, despite significant advances in machine learning, efficient and effective continual learning algorithms have not been demonstrated in various realistic scenarios \citep{prabhu2020gdumb}. 

Real-world learning scenarios often involve a large set of tasks that an intelligent agent must master throughout its lifetime. These tasks are not only encountered sequentially but also exhibit a long-tail distribution in terms of their sizes, i.e. the amount of available training data, reflecting the uneven distribution of information in the real world \citep{zhang2023deep,liu2019large}. Existing continual learning algorithms are usually developed on a small number of tasks with uniform sizes \citep{douillard2021continuum,de2021continual}, which may not fully capture some of these challenges. In addition, existing works \citep{gem,ewc,schwarz2018progress,derpp,dualnet,icarl,lwf} on such settings usually focus on smaller models trained from scratch, whereas pretrained models are becoming the go-to standard for numerous real-world applications \citep{bert,gpt3,vit}. A long-tail task sequence setting may be common for tasks involving pretrained models, given that a substantial portion of knowledge has already been acquired during the pre-training phase; consequently, subsequent updates might be frequent yet incremental in nature. For example, a large language model may have acquired good language skills during pretraining, but can still acquire new knowledge from daily news.

In this paper, we study the performance of continual learning on a large number of tasks that are drawn from long-tail task distributions. We do that by designing one synthetic dataset and two real-world continual learning datasets with these properties and studying the performance of various existing algorithms on them. In continual learning algorithms, the model parameters from a previous task usually serve as the initialization for learning a new task. In addition to that, information from previous learning sessions is usually used to mitigate forgetting while learning a new task. Typically, this is done by maintaining a replay buffer containing a subset of training data from previous tasks \citep{er, derpp, gem, agem}, or through regularization using statistics derived from earlier tasks \citep{ewc,ewcpp,si,mas}. 
% Another family of continual learning methods dynamically adjust model's architecture, allocates more parameters when learning a new task. But for a task sequence with numerous number of tasks, 

State of the art learning algorithms typically maintain some statistics along the learning procedure as its states. For example, the Adam optimizer \citep{adam} maintains the first and second moments of the gradients. A natural question to ask is whether these optimizer states can also be exploited to reduce catastrophic forgetting \citep{french1999catastrophic} in continual learning\footnote{We examined github repositories of multiple existing continual learning algorithms and found that existing algorithms would maintain the model parameters but reset all the other optimizer states at the start of a new task.}. We develop a method based on the Adam optimizer and show that utilizing the optimizer states can be effective for reducing forgetting, particularly for continual learning with a large number of tasks from long-tail task distributions. Indeed, simply maintaining the current moments of the gradient at the beginning of a new task, instead of resetting them, reduces forgetting in our experiments. However, moments of previous tasks will still be gradually forgotten because of the exponential moving average mechanism in Adam. To handle this, we keep a weighted average of the moments of all the tasks previously encountered, with task sizes being weights. The method is compatible with most existing continual learning methods with only a small amount of additional computation or memory cost. Experimentally, it is able to further reduce forgetting beyond what is already achieved by existing algorithms.  %%%%

%To address this limitation, we propose a new set of benchmarks that feature a long task sequence and long-tailed distributed task sizes.

% Continual learning often deals with tasks, and the model parameters of the previous task is used to initialize the model when learning a new task. Various continual learning methods are proposed to leverage the information of previous learning sessions to mitigate forgetting when learning a new task. Such information can be a subset of previous data or a regularization calculated from previous statistics. In this paper, we argue that there is a readily available information of previous learning sessions that has been overlooked, the optimizer states. We propose an algorithm to leverage the optimizer states to mitigate forgetting. The proposed algorithm is applicable on top of most existing continual learning methods with no additional computation or memory cost.

\vspace{-5pt}
\section{Moment continual optimizer}

We focus on the Adam \citep{adam} family of algorithm, including Adam and AdamW \citep{adamw}. Our modification of Adam and AdamW into their continual versions are shown in Alg. \ref{alg:adam}, the modified algorithms will be referred as Continual Adam and Continual AdamW in the rest of the paper.  AdamW is the same as Adam except for the addition of the highlighted part. For convenience, we will only refer to Adam in our description, even though the modifications are applicable to both algorithms.

\begin{algorithm}[h]
\caption{Continual Adam and \colorbox{lime}{AdamW} Algorithm}
\label{alg:adam}
\begin{algorithmic}[1]
\REQUIRE Learning rate $\alpha$, $\beta_1$, $\beta_2$, $\beta_3$, $f(\theta)$, $\epsilon$, \colorbox{lime}{$\lambda$}
\STATE Initialize $t=0$, $m_0=0$, $v_0=0$, $v_c=0$, $c=0$ 
\FOR{task in $\mathcal{T}$}  
\STATE $t\leftarrow 0$; $m_0\leftarrow 0$; $v_0\leftarrow 0$
\WHILE{stopping criterion not met}
\STATE $t \leftarrow t+1$
\STATE $g_t \leftarrow \nabla_{\theta}f_t(\theta_{t-1})$ \label{Cont:grad}
\STATE $m_t \leftarrow \beta_1 m_{t-1} + (1-\beta_1)g_t$  \COMMENT{First moment}\label{Cont:firstMoment}
\STATE $v_t \leftarrow \beta_2 v_{t-1} + (1-\beta_2)g_t^2$ \COMMENT{Second moment}\label{Cont:secondMoment}
\STATE $\hat{m}_t \leftarrow m_t/(1-\beta_1^t)$ \label{Cont:firstScale}
\STATE $\hat{v}_t \leftarrow v_t/(1-\beta_2^t)$ \label{Cont:secondScale}
\STATE $v_t' \leftarrow (t\hat{v}_t+cv_c)/(t+c)$ \label{Cont:mix}
\STATE $\theta_t \leftarrow \theta_{t-1} - (1-\beta_3^t)\alpha\hat{m}_t/(\sqrt{v_t'}+\epsilon)$\colorbox{lime}{$-\lambda\theta_{t-1}$} \label{Cont:paramUpdate}
\ENDWHILE
\STATE $v_c\leftarrow (t\hat{v}_t+cv_c)/(t+c)$ \COMMENT{Weighted average of task moments}\label{Cont:momentUpdate}
\STATE $c \leftarrow c+t$
\ENDFOR
\end{algorithmic}
\end{algorithm}

In continual learning, when starting to learn on a new task, the model is initialized from the learned parameters from the previous task.   
%Likely due to existing works usually work on small image classification datasets, and choose the stateless vanilla SGD as the optimizer due to its better generalizability on image tasks \citep{wilson2017marginal}. When it comes to NLP tasks, adaptive optimizers like Adam or AdamW are often preferred to pretrain or fine-tune a language model. 
Adaptive optimizers usually maintain state variables during learning; in the case of Adam, this would be the first and second moments of gradients. Existing works usually reset the optimizer states, i.e. a new instance of the optimizer is initialized every time when a new task comes \citep{gem,cao2021continual,cossu2022continual,wang2022learning,ke2022continual,lamol,douillard2022dytox,madotto2020continual,mi2020continual}. In this section, we argue that the optimizer state, specifically, the second moment, is useful information for continual learning, and we propose a simple algorithm (Alg. \ref{alg:adam}) to maintain and use the second moment along the training sequence. 

The algorithm is given in Algorithm \ref{alg:adam}. In the algorithm, all operations on vectors are done component-wise; Adam is an adaptive gradient method where each component of the parameter vector is given a different learning rate that is adapted at each timestep. The inner loop of the algorithm is mostly the same as Adam. The gradient is computed in Line~\ref{Cont:grad}. It is then used to update the first and second moments of the gradient in Line~\ref{Cont:firstMoment} and Line~\ref{Cont:secondMoment} by doing a weighted average of the new and old moments. The moments get bias corrected in Line~\ref{Cont:firstScale} and Line~\ref{Cont:secondScale}. Different from Adam, the second moment is then mixed with the stored second moment from earlier tasks in Line~\ref{Cont:mix} and used in the updating of the learning rate in Line~\ref{Cont:paramUpdate} where the parameters are updated by adding the first moment, scaled by the learning rate. Another difference from Adam is that we further scale the step size according to the step count in current task in Line~\ref{Cont:paramUpdate}. This is further discussed in Section \ref{sec:step}.

\subsection{Keeping the second moment from previous tasks}
We provide some heuristic justifications for why the second moment may be important for continual learning. The second moment estimates the variance of the gradient. Similar to EWC \citep{ewc}, we can approximate the posterior of the parameter given previous data as a Gaussian, whose variance is given by the second moment, and mean given by the parameter learned from previous tasks. To maximize the posterior, EWC regularizes the parameter to that learned from previous tasks, with the diagonal Fisher Information, i.e. reciprocal of the variance, as the regularization weight. Similarly, our approach of keeping the second moment of the optimizer can also be seen as a regularization, and allocates smaller step sizes to components that have larger variances. Different from EWC, instead of explicitly calculating the regularization as a loss, and calculating the variance at the end of each task, we directly modify the second moments in the optimizer and maintain an online approximation of the variance.

% Components of the parameter vector whose second moments are large have large average squared gradients, possibly over many examples in the past, indicating that it is more important to past examples than components whose second moments are small. The learning rate for each parameter vector component in Line~\ref{Cont:paramUpdate} is  essentially divided by the square root of its second moment. So effectively, the learning rate will be small for components which have been important in the past, and large on components that have not been important. By minimizing updates to previously important components of the parameter vector and focusing utilization on unused parts of the parameter vector, we may correspondingly reduce forgetting, which would be helpful for continuous learning.  

The second moment can also be seen as an approximation of the diagonal of the empirical Fisher Information Matrix (FIM) \citep{adam}. This also resembles the behavior of the AdaGrad algorithm \citep{adagrad} for online convex optimization, where the empirical FIM is used as a regularization matrix and is updated online. The AdaGrad algorithm in online learning has good regret behavior \citep{ocobook}. A continual learning algorithm has similar goals as we want to find the best model parameters in hindsight, similar to the parameters found by multi-task learning on all the tasks. The similarity suggests that preserving second moments from previous tasks when transitioning to a new task may be beneficial in continual learning. %The Fisher Information Matrix is also used in continual learning methods that are based on regularization methods \citep{ewc,schwarz2018progress,ewcpp} and we discuss the similarities to Continual Adam when describing the regularization methods later.

\subsection{Task-wise averaged second moments}
One difference between Adam and AdaGrad is that AdaGrad uses the sum of squares of gradients to regularize each optimization step, while Adam uses an exponential moving average of squares of gradients and is updated at each step (Alg. \ref{alg:adam} line 8). In continual learning, we operate on data organized in tasks, hence different parameters may get activated in different tasks. If we simply maintain the second moment $v_t$ across tasks, it may decrease exponentially fast for those parameters that are not activated by the current task, and we will lose those previous information. To address this issue, we maintain a weighted average of the second moments from each task using task sizes as weights, updated at the end of each task (Alg. \ref{alg:adam} line~\ref{Cont:momentUpdate}). At each optimization step, we use the weighted average of $v_c$ and the second moment of the current task $\hat{v}_t$ as the denominator (Alg. \ref{alg:adam} line~\ref{Cont:mix}). This enables us to preserve the second moment information from all previous tasks, instead of disproportionately allocating higher weights to more recent tasks.

\subsection{Learning rate adjustment at the beginning of a task}
\label{sec:step}
Adaptive algorithms such as Adam tend to make large steps at the beginning of the optimization. If we look at the Adam algorithm, the first optimization step $\hat{m}_1/(\sqrt{\hat{v}_1}+\epsilon)=g_1/(\sqrt{g_1^2}+\epsilon)\approx \text{sign}(g_1)$ is simply a sign function of the gradient $g_1$, whose magnitude is likely larger than updates at later steps. While this should be fine for single-task optimization, allowing for initial exploration of the parameter space, it could result in increased forgetting in continual learning scenarios as the exploration may cause the parameters to deviate more significantly from those of previous tasks. To mitigate this issue, we adjust the step size by a multiplier $1-\beta_3^t$ where $t$ is the step count of the current task (Alg. \ref{alg:adam} line~\ref{Cont:paramUpdate}). We choose $\beta_3=0.9$ in our experiments so that the multiplier will quickly grow from $0.1$ to $1$ and only affects the beginning of the optimization of a task. 
\vspace{-5pt}
\section{Existing methods}
\label{sec:methods}
We discuss some existing continual learning methods and its relationship to Continual Adam. These methods are also used in our experimental investigation.

\subsection{Regularization-based methods}
\paragraph{EWC} EWC \citep{ewc} is a regularization-based method, which regularizes model parameters to prevent them from moving too far from parameters learned from previous tasks. The regularizer is derived from the posterior of the parameter distributions given data from previous tasks, where each posterior is approximated by a Gaussian with the previous parameter as mean, and the diagonal of the empirical Fisher Information Matrix as the precision matrix. The loss function when learning task $\tau$ is given as 
\begin{equation}
    \label{equ:ewc}
    \mathcal{L}_\tau(\theta) = \mathcal{L}_{\mathcal{D}_\tau}(\theta) + \alpha\sum_{i=1}^{\tau-1} \|F_i(\theta-\theta^{i*})\|^2
\end{equation}
where $\mathcal{L}_{\mathcal{D}_\tau}(\theta)$ is the loss on the training data of task $\tau$, $\theta^{i*}$ is the learned parameters at the end of task $i$. $F_i$ is the diagonal matrix approximation of FIM at $\theta^{i*}$ on $\mathcal{D}_i$, which is evaluated at the end of task $i$. We note that EWC can be implemented efficiently without storing $F_i$ and $\theta^{i*}$ for each task. This has been alluded to previously, e.g. in \citep{schwarz2018progress}, without the details and implementation that we provide here. To calculate the gradient of Equ. \ref{equ:ewc}, we only need to maintain two variables while learning on the task sequence: $\sum_{i=1}^\tau F_i$ and $\sum_{i=1}^\tau F_i \theta^{i*}$. However, EWC does require an additional pass over the dataset $\mathcal{D}_i$ to calculate $F_i$ after learning each task $i$.

\paragraph{EWC++} EWC++ \citep{ewcpp} is proposed as a more efficient version of EWC. It maintains an approximation of the diagonal of the empirical Fisher Information Matrix, which is updated after every batch of data as $F^b=\alpha F'^b + (1-\alpha) F^{b-1}$, where $F'^b$ is calculated on the current batch. This avoids the additional pass of the current dataset at the end of each task. Moreover, it only regularizes to value of the parameters at the end of the last task instead regularizing it with parameters from every previous task as done in EWC. The loss function when learning task $\tau$ is given as
\begin{equation}
    \mathcal{L}_\tau(\theta) = \mathcal{L}_{\mathcal{D}_\tau}(\theta) + \alpha\|(F_{\tau-1}(\theta-\theta^{(\tau-1)*})\|^2
\end{equation}

Continual Adam that we propose is related to both EWC and EWC++. Consider the single task setting. Given the loss of all the data seen so far up to step $t$,  $\mathcal{L}_{\mathcal{D}_t}(\theta)$, and the estimated diagonal second moment matrix $F$ up to step $t$, the optimal solution of $\mathcal{L}_{\mathcal{D}_t}(\theta)+\|F(\theta - \theta^*)\|^2$ can be found by setting the gradient to zero, giving $\theta = \theta^*-F^{-1}\nabla \mathcal{L}_{\mathcal{D}_t}(\theta)$, which provides an Adam type update. However, Adam \citep{adam} use the inverse square root of the empirical Fisher Information Matrix $F^{-1/2}$ instead of $F^{-1}$; this appears to have better behaviour in optimization. The second moment maintained by Continual Adam can be viewed as an approximation to $\sum_{i=1}^\tau F_i$, which is maintained by EWC. As in EWC++, the regularization of Continual Adam constrains it to be near the solution of the previous task (the initial parameter), instead of near all the previous tasks as in EWC. Moreover, Continual Adam can be combined with EWC or EWC++ loss, as Continual Adam modifies the optimization process, while EWC and EWC++ modify the loss function.

\subsection{Replay-based methods}
Replay-based methods maintain a buffer containing a subset of training examples from previous tasks that is used alongside data from the current task to reduce the effect of forgetting. 

\paragraph{Reservoir} In the reservoir method \citep{reservoir,derpp}, a fix size buffer $\mathcal{M}$ is used to store a subset of previous data using the reservoir sampling algorithm, which has the property that all encountered examples appear in the buffer with equal probability. When learning on the new task, for each batch of new data, a batch of data is randomly sampled from the buffer, and one gradient optimization step is performed on the mixture of the two batches.

\paragraph{DER++} DER++ \citep{derpp} proposes to do knowledge distillation from the old model to the new model using a buffer. DER++ also uses a buffer with reservoir sampling and stores the model logits into the buffer. When learning a new task, in addition to the loss calculated on the buffer data, the model is also regularized to match the old model's logits on old data stored in the buffer. 

% Let $h_\theta(x)$ denote the logits of the model, and $f_\theta(x)$ denote the prediction of the model, the loss function when learning task $\tau$ is given as 
% \begin{equation}
%     \mathcal{L}_\tau(\theta) = \mathcal{L}_{\mathcal{D}_\tau}(\theta) + \mathbb{E}_{(x,y,z)\sim \mathcal{M}}[\|z-h_\theta(x)\|^2_2] + \mathbb{E}_{(x,y,z)\sim \mathcal{M}}[l(y, f_\theta(x)]
% \end{equation}

\paragraph{A-GEM} A-GEM \citep{agem} is an efficient version of GEM \citep{gem}. It maintains a fix size buffer to store data from previous tasks and allocate an equal amount of space for all tasks. The learning of the current task is then seen as a constrained optimization problem, where the constraint is that the average loss on the buffer should not increase.

For some tasks, having a buffer of training examples from previous tasks can be more effective in reducing forgetting than using regularization based methods. As discussed earlier, Continual Adam provides some of the effects of regularization based methods and can be used in conjunction with these replay-based methods to provide the advantages of both regularization and well as replay-based methods in reducing forgetting.

% \begin{algorithm}[H]
% \caption{Continual Adam Algorithm}
% \label{alg:adam}
% \begin{algorithmic}[1]
% \REQUIRE Learning rate $\alpha$, $\beta_1$, $\beta_2$, $f(\theta)$, $\epsilon$
% \STATE Initialize $t=0$, $m_0=0$, $v_0=0$, $k=0$, $v_c=0$
% \FOR{task in $\mathcal{T}$}
% \STATE $k\leftarrow k+1$; $t\leftarrow 0$; $m_0\leftarrow 0$; $v_0\leftarrow 0$
% \WHILE{stopping criterion not met}
% \STATE $t \leftarrow t+1$
% \STATE $g_t \leftarrow \nabla_{\theta}f_t(\theta_{t-1})$
% \STATE $m_t \leftarrow \beta_1 m_{t-1} + (1-\beta_1)g_t$
% \STATE $v_t \leftarrow \beta_2 v_{t-1} + (1-\beta_2)g_t^2$
% \STATE $\hat{m}_t \leftarrow m_t/(1-\beta_1^t)$
% \STATE $v'_t \leftarrow (1/k)*v_t+(1-1/k)*v_c$
% \STATE $\hat{v}_t \leftarrow v'_t/(1-\beta_2^t)$
% \STATE $\theta_t \leftarrow \theta_{t-1} - \alpha\hat{m}_t/(\sqrt{\hat{v}_t}+\epsilon)$
% \ENDWHILE
% \STATE $v_c\leftarrow (1/k)*v_t+(1-1/k)*v_c$
% \ENDFOR
% \end{algorithmic}
% \end{algorithm}
\vspace{-5pt}
\section{Datasets}
\subsection{Synthetic continual learning task}
We design a synthetic continual linear regression task sequence $\mathcal{T}$ to investigate how different factors affect continual learning performance. For each task $\tau$ in $\mathcal{T}$, we randomly initialize a $d$-dimensional weight vector $\bm{w}_\tau$ of the linear function, and a multi-layer perceptron (MLP) $g_\tau: \mathbb{R}^l\to\mathbb{R}^d$ which is used to generate the input by first generating a $l$-dimensional latent vector $z$ from a Gaussian distribution and transforming it into the input vector through $\bm{x}=g_\tau(\bm{z})$. The corresponding target value is given by $y=\bm{w}_\tau^{\top}\bm{x}+n$, where $n$ is noise sampled from a Gaussian distribution, and $\bm{w}_\tau$ represents the optimal parameter of the task. The continual learning algorithm needs to continually learn a model $f_{\bm{\theta}}:\mathbb{R}^d\to\mathbb{R}$ parameterized by $\bm{\theta}$ on the sequence of linear regression tasks generated using the process described above. 
%Given an input $\bm{x}$, the output of the model is given as $f_{\bm{\theta}}(\bm{x})=\bm{\theta}^{\top}\bm{x}$.

In this synthetic task, we can easily control the total number of tasks in $\mathcal{T}$ as well as the number of examples in each task, allowing us to generate a large number of tasks from a long tailed distribution of task sizes. We can also control the similarities of the tasks by controlling $\bm{w}_\tau$ allowing us to examine the effects of different assumptions of task similarities. Furthermore, in this setting, each $g_\tau$ represents a different data distribution, where each task has data coming from different underlying distributions, resembling the problem of domain-incremental continual learning. 

% We try to answer three questions by experimenting with the synthetic task: 1) How does the length of the task sequence affect continual learning performance? 2) How do task sizes and orders affect continual learning performance? 3) How does Moment-continual Optimizer affect continual learning performance in different task sequences? We continually learn a model $f_{\bm{\theta}}:\mathbb{R}^d\to\mathbb{R}$ parameterized by $\bm{\theta}$ on the sequence of linear regression tasks. Given an input $\bm{x}$, the output of the model is given as $f_{\bm{\theta}}(\bm{x})=\bm{\theta}^{\top}\bm{x}$.

%We also propose two continual problems using real-world data. The two datasets also feature a long-tail task size distribution, which is similar to human learning, where a large chunk of the knowledge is learned at an early stage.

\subsection{Continual word sense disambiguation}
We repurpose Word Sense Disambiguation (WSD) \citep{navigli2009word} as a continual learning problem (\textbf{WSD-CL}). Word sense disambiguation is a classic natural language processing task where we would have a target word and a target sentence as input, and the goal is to disambiguate the target word in the sentence by giving its correct sense in WordNet \citep{wordnet}. For example, given a target sentence \textit{"A cat is chasing a mouse."} and the target word \textit{"mouse"}, we should select the sense \textit{"any of numerous small rodents..."} from the WordNet senses of the word \textit{"mouse"}. We propose to treat each target word as a task, and the corresponding target sentences containing the word as data for the task. Since the usage of words follows a long-tail distribution, the sizes of our task also have a natural long-tail distribution. We order the task by sampling without replacement, with the selection probability based on the task size, which results in a task sequence where larger tasks appear more frequently at the early stage.

We use the SemCor 3.0 corpus to construct our continual learning sequence, which results in 20399 tasks. For each task, we randomly select 90\% of the target sentences as training data, and the remaining 10\% as testing data. For tasks with less than 2 target sentences, we use all of them as testing data. These tasks can be viewed as zero-shot tasks within the sequence.

\subsection{Continual visual question answering}
Similar to \textbf{WSD-CL}, we construct another continual learning problem that has a visual component. We repurpose the visual question answering (VQA) \citep{vqa} problem as a continual learning problem (\textbf{VQA-CL}). The visual question answering problem takes an image and a natural language question pair as input, where the answer to the question, e.g. \textit{"What is the color of the bottle on the table?"}, can be found in the corresponding image. The question often requires the ability to correctly recognize different objects in the image and to reason about their spatial relationship or their attributes, such as color. We treat each object type as a task and use the questions that contain the object and corresponding images as training data for the task. For example, the \textit{"table"} task would be question-image pairs where the questions contain the word \textit{"table"}. To learn on this task sequence, the model will need to continually learn new objects as described in natural language as well as their appearances. This can be seen as a form of multi-modal continual learning. Note that such a split of tasks would have overlaps between tasks, for example, the above question would be in both the \textit{"table"} task and the \textit{"bottle"} task. While different from existing continual learning formulations which often involve separate, non-overlapping tasks, this formulation resembles real-world situations where tasks may share common elements.

We use the GQA dataset \citep{gqa} to construct our continual learning sequence. The GQA dataset features a scene graph for each image, and the questions are automatically generated using natural language templates and the scene graph. Questions generated this way require more understanding of a particular object, making it suitable for our object-level continual learning problem. We select 1000 objects to formulate 1000 tasks and order them by how frequently the object appears in the whole dataset, which results in a natural long-tail distribution of tasks.

\vspace{-5pt}
\section{Experiments}

In this section, we first use the synthetic dataset to study the effect of first moment, second moment, and learning rate adjustment in a long-tail task sequence. Then we experiment on our proposed long-tail task sequence \textbf{WSD-CL} and \textbf{VQA-CL}. We test existing continual learning works in such a setting, and we show that existing methods can gain further improvements by using our proposed method.

\subsection{Evaluation metric}
To evaluate the continual learning performance of a model, we use four metrics: Retained Performance (RP), Learning Performance (LP), Backward Transfer (BWT), and Forgetting (FGT) \citep{gem,ewcpp}. Let $T$ denote the number of all tasks, $n_i$ denote the number of testing examples in task $i$, $a_{i,j}$ denote the score of the model on task $j$ after learning on task $i$. The score depends on the evaluation metric of the task itself, which is accuracy for classification tasks, and mean square error for the synthetic linear regression problem.
\begin{equation}
\begin{aligned}
\text { RP } & : \frac{1}{\sum_{i=1}^T n_i} \sum_{i=1}^T n_i a_{T, i} & \text { BWT } & :\frac{1}{\sum_{i=1}^{T-1} n_i} \sum_{i=1}^{T-1} n_i(a_{T, i}-a_{i, i}) \\
\text { LP } & : \frac{1}{\sum_{i=1}^T n_i} \sum_{i=1}^T n_i a_{i, i} & \text { FGT } & :\frac{1}{\sum_{i=1}^{T-1} n_i} \sum_{i=1}^{T-1} n_i(\max_{j \in [i,T]} a_{j,i} - a_{T, i})
\end{aligned}    
\end{equation}

The four metrics measure different aspects of continual learning. RP measures the average performance at the end of the sequence; BWT measures how future tasks improve the performance of previously learned tasks; LP measures the plasticity of the model; FGT measures the amount of forgetting during the learning sequence. Note that the forgetting metric (FGT) has a term $\max_{j\in [i,T]}a_{j,i}$, assuming the metric of the task is higher better, otherwise the term should be \texttt{min}. Moreover, this term requires evaluating the model on all seen tasks after training on each task. The computation of this evaluation is quadratic to the number of tasks. When dealing with a small number of tasks, the computation is manageable. But for problems with thousands of tasks, the evaluation becomes too burdensome. Thus, instead of evaluating after learning every task, we evaluate the model after every $k$ tasks, which offers a summary of performance changes for each task during the learning process.

\subsection{Controlled study on synthetic dataset}

\begin{table}[htbp]
\caption{Ablation study on the synthetic dataset. "EA" represents exponential moving average, i.e. simply keep the moments in optimizer states (instead of initializing the moving average as zero, use the optimizer states from the previous task to initialize). "Reset" represents resetting the optimizer state to zeros. "TA" represents task-wise average, as in Alg. \ref{alg:adam}. Numbers reported are average of 100 runs. Metric FGT and BWT are reported in the appendix.}
%SUPPLEMENT
\label{tab:syn_ablation}
\centering
\begin{tabular}{lccccccccc}
\hline
   & \multicolumn{1}{l}{\multirow{2}{*}{\begin{tabular}[c]{@{}c@{}}1st\\ moment\end{tabular}}} & \multirow{2}{*}{\begin{tabular}[c]{@{}c@{}}2nd\\ moment\end{tabular}} & \multirow{2}{*}{\begin{tabular}[c]{@{}c@{}}lr\\ adjust\end{tabular}} & \multicolumn{3}{c}{RP$\downarrow$}                         & \multicolumn{3}{c}{LP$\downarrow$} \\
\#  & \multicolumn{1}{l}{}                                                                      &                                                                       &                    & same          & perturb       & shift          & same & perturb & shift \\ \hline
1  & Reset                                                                                     & Reset                                                                 & \cmark               & 0.65          & 1.12          & 18.20          & 0.01 & 0.01    & 0.01  \\
2  & Reset                                                                                     & Reset                                                                 & \xmark               & 2.19          & 2.93          & 23.45          & 0.01 & 0.01    & 0.01  \\ \hline
3  & Reset                                                                                     & EA                                                                    & \cmark               & 0.13          & 0.44          & 17.81          & 0.01 & 0.01    & 0.01  \\
4  & Reset                                                                                     & EA                                                                    & \xmark               & 0.15          & 0.48          & 17.77 & 0.01 & 0.01    & 0.01  \\
5  & EA                                                                                        & EA                                                                    & \cmark               & 1.77          & 2.06          & 18.18          & 0.01 & 0.01    & 0.02  \\
6  & EA                                                                                        & EA                                                                    & \xmark               & 29.80         & 29.48         & 33.87          & 0.05 & 0.05    & 0.06  \\ \hline
7  & Reset                                                                                     & TA                                                                    & \cmark               & \textbf{0.10} & \textbf{0.43} & 18.08          & 0.01 & 0.01    & 0.01  \\
8  & Reset                                                                                     & TA                                                                    & \xmark               & 0.14          & 0.47          & \textbf{17.73}          & 0.01 & 0.01    & 0.01  \\
9  & TA                                                                                        & TA                                                                    & \cmark               & 85.96         & 86.74         & 107.70         & 1.09 & 1.03    & 0.86  \\
10 & TA                                                                                        & TA                                                                    & \xmark               & 377.87        & 378.97        & 408.48         & 2.82 & 2.71    & 1.83  \\ \hline
11 & \multicolumn{3}{c}{MTL}                                                                                                                                                                  & 0.18          & 0.39          & 6.28           & -    & -       & -     \\ \hline
\end{tabular}
\end{table}

%The synthetic dataset allows us to control the number of tasks as well as the number of examples in each task. 
To study continual learning on a long-tail distributed task sequence, we generate 1000 tasks with task sizes randomly drawn from a power distribution. The tasks are ordered by their sizes from large to small. %Such a task sequence is different from existing benchmarks and allows us to investigate how would continual learning work in such a setting.

The synthetic dataset allows us to control how diverse the tasks are by controlling $\bm{w}_t$. We motivate three settings: 1) \texttt{same}: all tasks have the same $\bm{w}_t$, which implies there is no interference between tasks, and there exists a "perfect" model that can do well on all tasks, although the input distribution will change with tasks. 2) \texttt{perturb}: we first choose $\bm{w}_1$ randomly, then $\bm{w}_2, \bm{w}_3, ...$ are generated by randomly choosing $k$ dimensions of $\bm{w}_1$, and replace it with random values. This setting resembles the pretrain-finetune paradigm, where we have one pretrained model (in this case $\bm{w}_1$), and for each task, the optimal parameter can be obtained by making modifications to $\bm{w}_1$. 3) \texttt{shift}: we obtain $\bm{w}_t$ by perturbing $k$ dimensions of $\bm{w}_{t-1}$ instead of $\bm{w}_1$. This is a more challenging setting as the optimal parameters of the tasks are slowly changing and the tasks interfere with each other. In such a setting, whether a linear regression model is suitable to achieve good performance on all tasks becomes questionable, which also resembles real-world situations where model capacity may become insufficient as the number of tasks grows.
We experiment with the Continual Adam algorithm (Alg. \ref{alg:adam}) in these scenarios and investigate three questions using the synthetic dataset: 
\paragraph{How to reuse the moments from previous tasks?} The simplest way is just to keep the optimizer states instead of initializing them as zeros when starting a new task. For Adam optimizer, it means we are maintaining an exponential moving average of the moments (EA). Another way is to maintain a copy of the moments and update it at the end of each task, which is a task-wise average of the moments (TA), as in Alg. \ref{alg:adam}. From Table \ref{tab:syn_ablation} we see that keeping second moments with TA shows better performance than EA in \texttt{same} and \texttt{perturb} settings, and shows similar performance in \texttt{shift} setting. Moreover, we see that both methods perform better than resetting the first and second moments (row 2 and 3), which indicates the effectiveness of keeping the second moments.
% Comparing row 3 with row 7 and row 4 with row 8 in Table \ref{tab:syn_ablation}, we see that it is desirable to use TA to keep second moments in both \texttt{same} and \texttt{perturb} setting, while in \texttt{shift} setting, TA and EA have similar performance. %TODO

\paragraph{Are first moments also useful to keep?} Given the good performance of keeping second moments, a natural question to ask is whether the first moments are also useful to keep. We experiment with keeping the first moment using either EA or TA. From Table \ref{tab:syn_ablation} we see that keeping the first moment is often not as good as only keeping the second moment. The LP of keeping the first moments is significantly worse than other methods. This is likely because keeping the first moments gives inertia to gradients and creates an optimization direction that is influenced by both historical and current task gradients, which leads to interference in the current task's learning process.
%, likely because keeping the first moments is equivalent to using an average of the history the first moment and the current gradient as the update direction, which will cause the optimization of the current task to be predominantly influenced by the first moments stored from previous tasks, instead of the gradient on the current task.

\paragraph{How effective is the learning rate adjustment?} As discussed in Sec. \ref{sec:step}, we restrict the step size at the beginning of the task to reduce exploration. From results in Table. \ref{tab:syn_ablation} we see that when using such adjustment, the LP metric is not affected, indicating the plasticity of the model is not harmed, while the RP metric is better. Interestingly, when we keep both the first moments and the second moments, methods with learning rate adjustment even show better LP than those without it. This observation further indicates that the first moments of previous tasks may have a negative impact on the learning of the current task. By using a small learning rate at the beginning of a task, we effectively limit the influence of historical moments in the current task's optimization process. Because the optimization steps at the beginning of a task for both EA and TA are predominantly computed using the historical moments, given that the weights of current task moments are small.

% In all three scenarios, our proposed algorithm (Table \ref{tab:syn_ablation}, line 7) shows better or comparable performance compared to other variants.

% \begin{table}[htbp]
% \caption{Synthetic experiments on 1000 long-tail distributed tasks. $\bm{w}$ is same for all tasks.}
% \label{tab:syn_lt}
% \centering
% \begin{tabular}{ccccccccc}
% \hline
% \multicolumn{1}{l}{} & \multicolumn{2}{c}{RP$\downarrow$}   & \multicolumn{2}{c}{LP$\downarrow$} & \multicolumn{2}{c}{FGT$\uparrow$} & \multicolumn{2}{c}{BWT$\downarrow$} \\
% method               & w/ MC      & w/o MC      & w/ MC     & w/o MC     & w/ MC      & w/o MC     & w/ MC      & w/o MC     \\ \hline
% Finetune             & 0.31       & 1.73        & 0.01      & 0.01       & -0.37      & -2.19      & 0.30       & 1.72       \\
% EWC                  & 0.31       & 2.10        & 0.01      & 0.01       & -0.37      & -2.69      & 0.30       & 2.10       \\
% EWC++                & 0.30       & 0.90        & 0.01      & 0.01       & -0.35      & -1.09      & 0.29       & 0.89       \\
% Reservoir            & 0.11       & 0.68        & 0.01      & 0.01       & -0.14      & -0.82      & 0.10       & 0.68       \\
% A-GEM                & 15.70      & 28.95       & 0.03      & 0.05       & -29.55     & -51.26     & 15.67      & 28.91      \\
% MTL                  & \multicolumn{2}{c}{0.18} & -         & -          & -          & -          & -          & -          \\ \hline
% \end{tabular}
% \end{table}
\vspace{-5pt}
\subsection{Experiments on real-world long-tail task sequences}
We test the methods we discussed in Sec. \ref{sec:methods} on our proposed datasets, two task sequences with long-tail task sizes, \textbf{WSD-CL} and \textbf{VQA-CL}. Apart from above mentioned methods, we also experiment with: 1) Finetune: simply train the model with training data of current task with no regularization or replay. 2) Multi-task Learning (MTL): Train with the mixture of training data of all tasks. Often seen as an upper bound for continual learning. We will show that our proposed algorithm can be combined with continual methods and yield further improvement. 

We use pretrained models as backbone for both dataset. For \textbf{WSD-CL}, we formulate the word sense disambiguation task as sentence matching between the target sentence and the senses of the target word \citep{glossbert}. We use BERT \citep{bert} as the backbone model and use an MLP binary classifier on the \texttt{[CLS]} token to decide a matching score for sentence pairs. The sense with the highest matching score is selected as the model's prediction. This formulation avoids using task-specific classification heads, which allows the model to have a constant number of parameters. For \textbf{VQA-CL}, we use Oscar \citep{oscar} as the backbone, which is a pretrained multi-modal transformer that takes both text tokens and image tokens as input. We apply an MLP classifier on the \texttt{[CLS]} token and select one answer from all possible answers. For each task in the task sequence, we allocate 5\% of the training data as a validation set. We train 4 epochs for each task and select the best performing model on the validation set. We use AdamW and Continual AdamW for both tasks.

\begin{table}[htbp]
\caption{Results on \textbf{WSD-CL}. "w/MC" columns are results of methods extended with our proposed algorithm, "w/o MC" columns are results of methods that reset optimizer states for each task.}
\label{tab:wsd_lt}
\centering
\begin{tabular}{ccccccccc}
\hline
\multicolumn{1}{l}{} & \multicolumn{2}{c}{RP$\uparrow$}    & \multicolumn{2}{c}{LP$\uparrow$}                                 & \multicolumn{2}{c}{FGT$\downarrow$}                                & \multicolumn{2}{c}{BWT$\uparrow$}                                \\
Method & w/ MC       & w/o MC      & \multicolumn{1}{l}{w/ MC} & \multicolumn{1}{l}{w/o MC} & \multicolumn{1}{l}{w/ MC} & \multicolumn{1}{l}{w/o MC} & \multicolumn{1}{l}{w/ MC} & \multicolumn{1}{l}{w/o MC} \\ \hline
Finetune             & 78.76       & 61.63       & 81.27                     & 79.53                      & 5.50                      & 25.98                      & -2.51                     & -17.90                     \\
EWC                  & 78.32       & 72.48       & 81.11                     & 79.56                      & 6.00                      & 10.65                      & -2.78                     & -7.08                      \\
EWC++                & 78.10       & 70.99       & 80.61                     & 75.48                      & 5.47                      & 8.88                       & -2.52                     & -4.49                      \\
Reservoir            & 79.78       & 74.48       & 80.82                     & 78.84                      & 3.91                      & 9.15                       & -1.04                     & -4.36                      \\
DER++                & 79.12       & 78.32       & 79.48                     & 79.41                      & 4.02                      & 4.82                       & -0.36                     & -1.10                      \\
AGEM                 & 78.95       & 69.37       & 81.22                     & 79.56                      & 5.17                      & 16.50                      & -2.27                     & -10.19                     \\
MTL                  & \multicolumn{2}{c}{83.72} & \multicolumn{1}{l}{-}     & \multicolumn{1}{l}{-}      & \multicolumn{1}{l}{-}     & \multicolumn{1}{l}{-}      & \multicolumn{1}{l}{-}     & \multicolumn{1}{l}{-}      \\ \hline
\end{tabular}
\end{table}
\vspace{-5pt}
\begin{table}[htbp]
\caption{Results on \textbf{VQA-CL}.}
\label{tab:gqa_lt}
\centering
\begin{tabular}{ccccccccc}
\hline
          & \multicolumn{2}{c}{RP$\uparrow$}   & \multicolumn{2}{c}{LP$\uparrow$} & \multicolumn{2}{c}{FGT$\downarrow$} & \multicolumn{2}{c}{BWT$\uparrow$} \\
Method          & w/ MC      & w/o MC      & w/ MC     & w/o MC     & w/ MC      & w/o MC     & w/ MC      & w/o MC     \\ \hline
Finetune  & 69.98      & 64.99       & 67.29     & 65.69      & 6.17       & 8.59       & 2.68       & -0.70      \\
EWC       & 69.62      & 67.27       & 66.16     & 66.16      & 5.85       & 7.84       & 3.46       & 1.11       \\
EWC++     & 69.40      & 64.07       & 66.79     & 65.29      & 6.49       & 9.15       & 2.61       & -1.22      \\
Reservoir & 71.61      & 67.44       & 68.51     & 67.05      & 5.51       & 7.45       & 3.09       & 0.39       \\
DER++     & 71.37      & 66.63       & 68.14     & 66.61      & 5.47       & 7.18       & 3.22       & 0.02       \\
AGEM      & 69.98      & 64.57       & 66.69     & 65.52      & 6.19       & 8.93       & 3.30       & -0.95      \\
MTL       & \multicolumn{2}{c}{73.98} & -         & -          & -          & -          & -          & -          \\ \hline
\end{tabular}
\end{table}

Results are shown in Table \ref{tab:wsd_lt} and \ref{tab:gqa_lt}. The continual learning methods mostly work better than simply transferring the parameters from the previous task to the new task (Finetune w/o MC). Replay-based methods mostly work better than regularization-based methods. For the regularization-based method, EWC mostly works better than EWC++. For the replay-based methods, Reservoir works quite well, although DER++ with distillation loss is more helpful in \textbf{WSD-CL}. We see that our proposed Continual AdamW consistently improves the performance over the original AdamW algorithm for both replay-based methods and regularization-based methods. Basic Continual AdamW (Finetune) mostly already outperforms all methods that do not use the second moment in the same way and using it on top of the replay-based methods can bring further improvements.  

We also see that in such a long-tail setting with pretrained models, the gap between different methods and MTL is less significant, especially when using the Continual AdamW algorithm. This is different from existing continual learning benchmarks, where the gap between different methods can be much larger. The observation suggests that pretrained models may suffer less from forgetting, and more efficient methods with less memory and computation requirement may be developed for pretrained models in such a setting.

%\subsection{Experiments on short, uniform task sequences}
We also investigate how our proposed algorithm performs on short, uniform task sequences. We experiment with two image classification continual learning tasks \citep{de2021continual} sequential-MNIST \citep{mnist} and sequential-CIFAR10 \citep{cifar10}, both of which are 5-task sequences with each task introducing two new classes. We also experiment with one language continual learning task \citep{lamol}, which consists of 5 NLP tasks formulated as question answering: SQuAD, WikiSQL, SST, QA-SRL, and WOZ. On sequential-MNIST and sequential-CIFAR10, Continual AdamW improves over AdamW in most methods, while in the NLP tasks, Continual AdamW has similar performance compared to AdamW. Since the 5 NLP tasks are quite different from each other, this may suggest that Continual Adam and Continual AdamW are more suitable for the scenario where tasks are similar to each other, as also indicated by Table \ref{tab:syn_ablation}. Experimental results are given in the appendix.

Our proposed approach of keeping a task-wise second moment in the optimizer is similar to EWC and EWC++ as they all leverage the empirical Fisher Information Matrix to formulate regularization. Experimentally, Continual Adam appears to be more effective. %The role of empirical Fisher Information Matrix in optimization is not fully understood \citep{limitationEF}.
Continual Adam uses the square root of the empirical Fisher Information Matrix, instead of the empirical Fisher Information Matrix. This subtle difference may cause the model to end up at different local minima and potentially impacts the generalization performance on unseen data \citep{wilson2017marginal}.  In our experiments, Continual Adam works better than EWC and EWC++ and combines well with replay-based methods, improving the performance of those methods. It would be useful to have a better theoretical understanding of Continual Adam and how it works in continual learning in future works.
\vspace{-5pt}
\section{Related works}
There are a few works on continual learning on scenarios with long tasks sequences. \textit{SCoLe} \cite{Lesort2022ScalingTN} works on continual learning with thousands of tasks. They follow the "finite world" assumption \citep{mundt2023wholistic} and allow overlap between tasks, and the task sizes are uniform. Another work also proposed a continual learning task sequence with thousands of tasks of uniform sizes \citep{wortsman2020supermasks}. This is achieved by randomly permuting the MNIST dataset, where each permutation serve as a task. In contrast, we do not enforce overlapping between tasks, and we work on long-tail task sizes with complicated real-world tasks, WSD and VQA, which encourage the use of pretrained models.

Some works suggested that the temporal misalignment between training and evaluating corpus may cause the model to have bad performance \citep{luu2021time,chen2021dataset}. Consequently, some studies have introduced benchmarks for continual learning of pretrained models, assessing their ability to learn and update factual knowledge by building training corpus from different time periods \citep{jang2021towards,jang2022temporalwiki}. Other works also proposes continual learning benchmark for continual learning by repurposing existing dataset \citep{lamol, DBLP:conf/iclr/WuCLLQH22, DBLP:conf/naacl/JinZZ00WA022,de2019episodic}, which typically feature a small number of tasks.

Recent works have proposed methods that focus on continual learning of pretrained models: MbPA++ \citep{de2019episodic} proposes to use replay and dynamic evaluation for continual learning of language models. LAMOL \citep{lamol} proposes to use the language model itself to generate data from old tasks and train on the mixture of the generated data and new data to mitigate forgetting. Some works also propose learning task-specific prompts for continual learning \citep{wang2022learning, wang2022s}.
\vspace{-5pt}
\section{Conclusion}
We study continual learning with numerous tasks from long-tail distributions, a natural setting in practice. We design two continual learning problems with such properties using word sense disambiguation and visual question answering, and experiment with existing continual learning algorithms in this setting. We propose a simple algorithm that modifies the Adam optimizer for continual learning, which uses the weighted average of the second moments from previous tasks together with Adam to reduce forgetting. The algorithm is shown effective in experiments, and it can be used in conjunction with existing continual learning algorithms to further improve performance.

\bibliography{bib_dblp}
\bibliographystyle{ieeetr}

\end{document}

% \maketitle
\appendix
\section{Experiments on short, uniform task sequences}
% So far we have seen that our proposed Continual Adam and AdamW algorithm can improve performance on long-tail task sequences, we are also interested in its effectiveness in existing continual learning benchmarks with short and uniform task sizes. 
We experiment with two image classification continual learning tasks \citep{de2021continual} sequential-MNIST \citep{mnist} and sequential-CIFAR10 \citep{cifar10}, both of which are 5-task sequences with each task introducing two new classes. We also experiment with one language continual learning task \citep{lamol}, which consists of 5 NLP tasks formulated as question answering: SQuAD, WikiSQL, SST, QA-SRL, and WOZ. We compare AdamW and Continual AdamW for all tasks.

\begin{table}[htbp]
\caption{Results on seq-MNIST and seq-CIFAR10. Numbers reported are average of 10 runs.}
\label{tab:image_classification}
%SUPPLEMENT
\centering
\begin{tabular}{ccccc}
\hline
\multicolumn{1}{l}{} & \multicolumn{2}{c}{seq-MNIST}               & \multicolumn{2}{c}{seq-CIFAR10} \\
Method & w/ MC                & w/o MC               & w/ MC          & w/o MC         \\ \hline
Finetune             & 19.11                & 18.79                & 19.58          & 19.56          \\
EWC                  & 44.94                & 41.29                & 19.30          & 19.41          \\
EWC++                & 19.78                & 19.75                & 19.64          & 19.63          \\
Reservoir            & 81.18                & 78.56                & 60.44          & 61.72          \\
DER++                & 87.12                & 85.32                & 64.73          & 62.93          \\
A-GEM                 & 71.42                & 70.08                & 25.09          & 21.56          \\ 
MTL & \multicolumn{2}{c}{95.64} & \multicolumn{2}{c}{91.17}\\
\hline
\end{tabular}
\end{table}

\begin{table}[htbp]
    \caption{Results of continual learning of language on a sequence of five tasks: SQuAD, WikiSQL, SST, QA-SRL, and WOZ. All tasks are formulated as question answering.}
    \label{tab:lamol}
    \centering
\begin{tabular}{lllllllll}
\hline
                              & \multicolumn{2}{c}{RP$\uparrow$}                       & \multicolumn{2}{c}{LP$\uparrow$}                       & \multicolumn{2}{c}{FGT$\downarrow$}                    & \multicolumn{2}{c}{BWT$\uparrow$}                      \\
Method                        & \multicolumn{1}{c}{w/ MC} & \multicolumn{1}{c}{w/o MC} & \multicolumn{1}{c}{w/ MC} & \multicolumn{1}{c}{w/o MC} & \multicolumn{1}{c}{w/ MC} & \multicolumn{1}{c}{w/o MC} & \multicolumn{1}{c}{w/ MC} & \multicolumn{1}{c}{w/o MC} \\ \hline
\multicolumn{1}{c}{Finetune}  & 36.88                     & 45.32                      & 77.64                     & 77.53                      & 40.76                     & 32.20                      & -40.76                    & -32.20                     \\
EWC                           & 57.74                     & 56.94                      & 75.46                     & 74.99                      & 24.35                     & 25.46                      & -17.72                    & -18.05                     \\
EWC++                         & 58.82                     & 60.99                      & 58.82                     & 75.17                      & 22.73                     & 21.60                      & -15.78                    & -14.18                     \\
\multicolumn{1}{c}{Reservoir} & 76.29                     & 75.61                      & 78.68                     & 75.61                      & 6.56                      & 6.96                       & -2.39                     & -2.47                      \\
DER++                         & 66.22                     & 67.46                      & 68.99                     & 69.96                      & 11.18                     & 9.86                       & -2.77                     & -2.50                      \\
A-GEM & 60.78 & 65.85 & 78.57 & 78.84 & 20.79 & 15.91 & -17.78 & -12.99\\ \hline
\end{tabular}
\end{table}

We see that the effect of Continual AdamW is mixed on these 5-task continual learning tasks. Continual AdamW consistently improve performance on sequential-MNIST dataset, while having mixed performance on sequential-CIFAR10 and the NLP task sequence.

\section{Effect of number of tasks}
\begin{figure}[htbp]

    \begin{subfigure}{.5\textwidth}
        \centering
        \includegraphics[width=\linewidth]{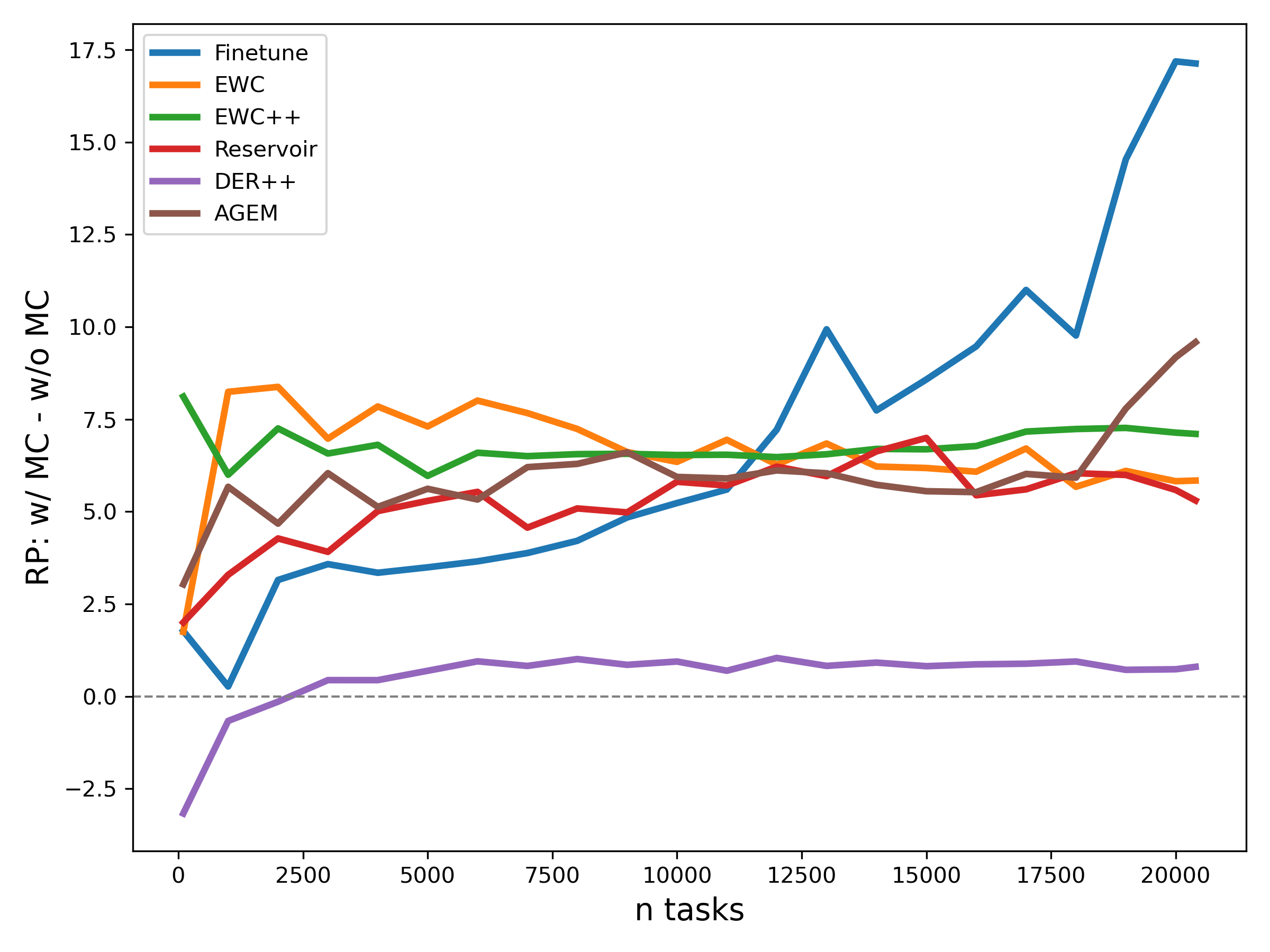}
        \caption{\textbf{WSD-CL}}
    \end{subfigure}
    \begin{subfigure}{.5\textwidth}
        \centering
        \includegraphics[width=\linewidth]{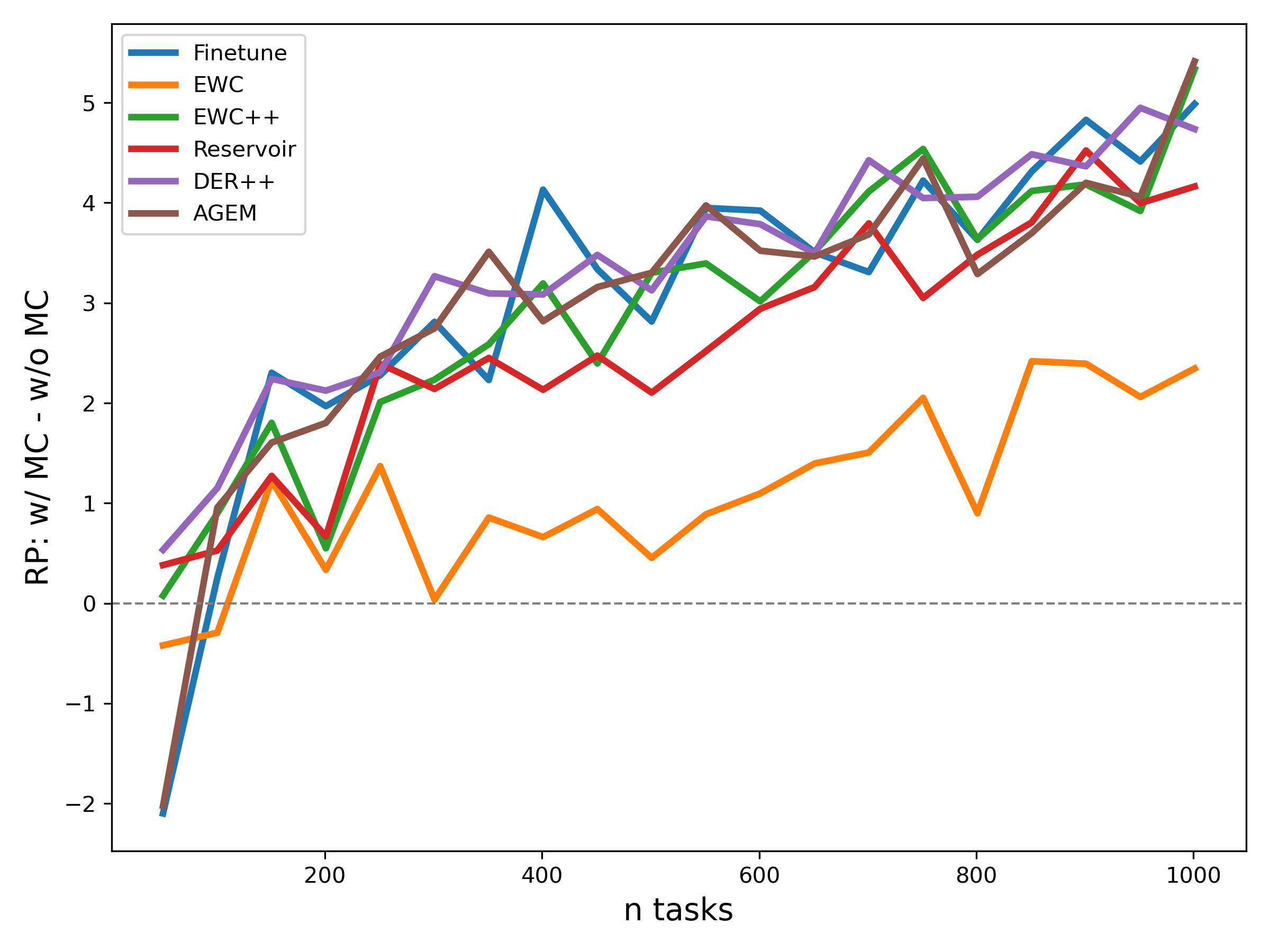}
        \caption{\textbf{VQA-CL}}
    \end{subfigure}
    \caption{Difference between Continual AdamW and AdamW on \textbf{WSD-CL} and \textbf{VQA-CL} by number of tasks. X axis is number of tasks; Y axis is the difference between Continual AdamW and AdamW on RP metric (greater than zero means Continual AdamW is better). For \textbf{WSD-CL}, model is evaluated at $100, 1000, 2000, 3000, ..., 20000, 20399$-th tasks. For \textbf{VQA-CL}, model is evaluated at $50, 100, 150, 200, ..., 1000$-th tasks.}
    \label{fig:real}
\end{figure}

\begin{figure}[htbp]

    \begin{subfigure}{.33\textwidth}
        \centering
        \includegraphics[width=\linewidth]{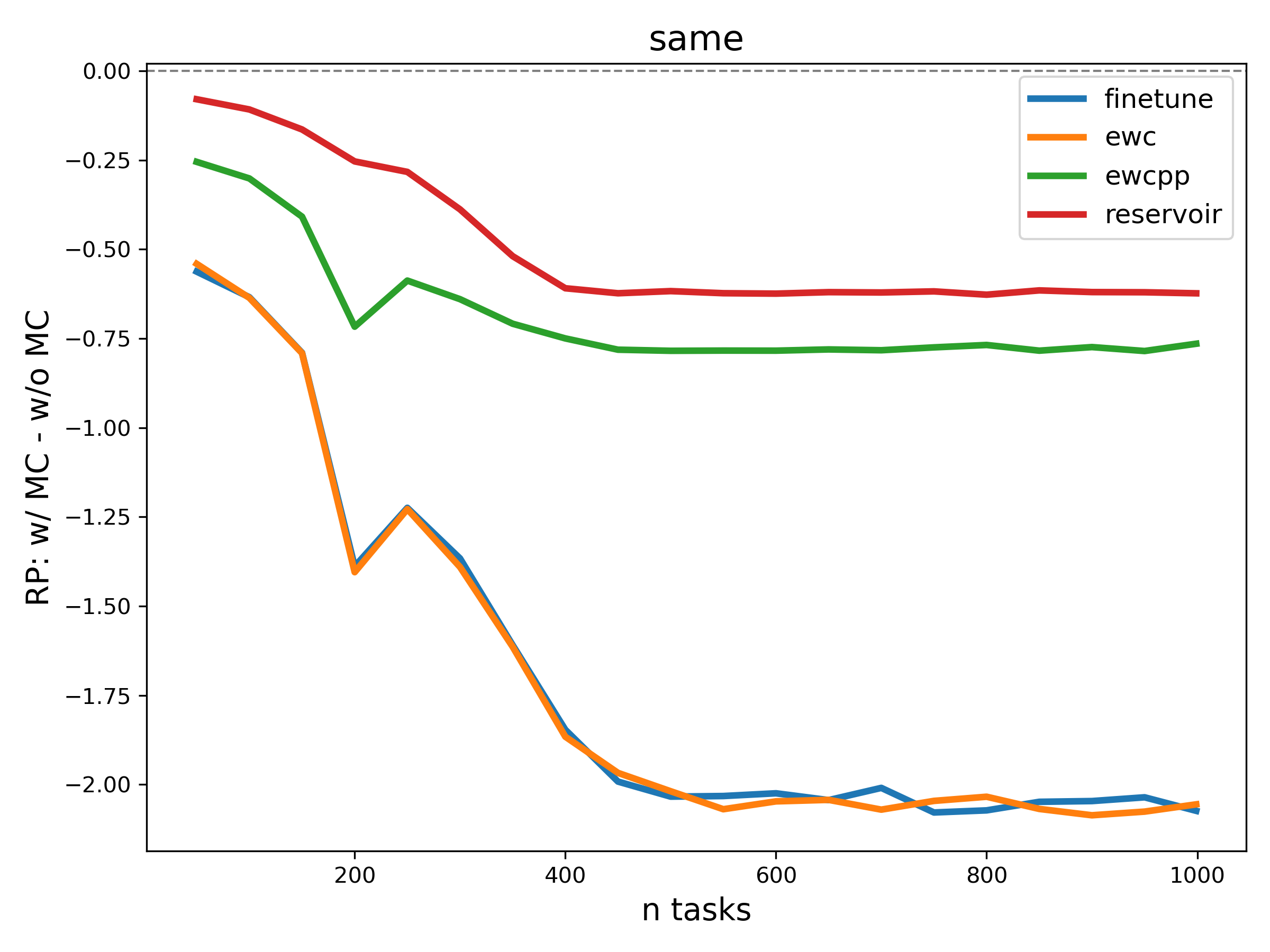}
        \caption{Synthetic task \texttt{same}.}
    \end{subfigure}
    \begin{subfigure}{.33\textwidth}
        \centering
        \includegraphics[width=\linewidth]{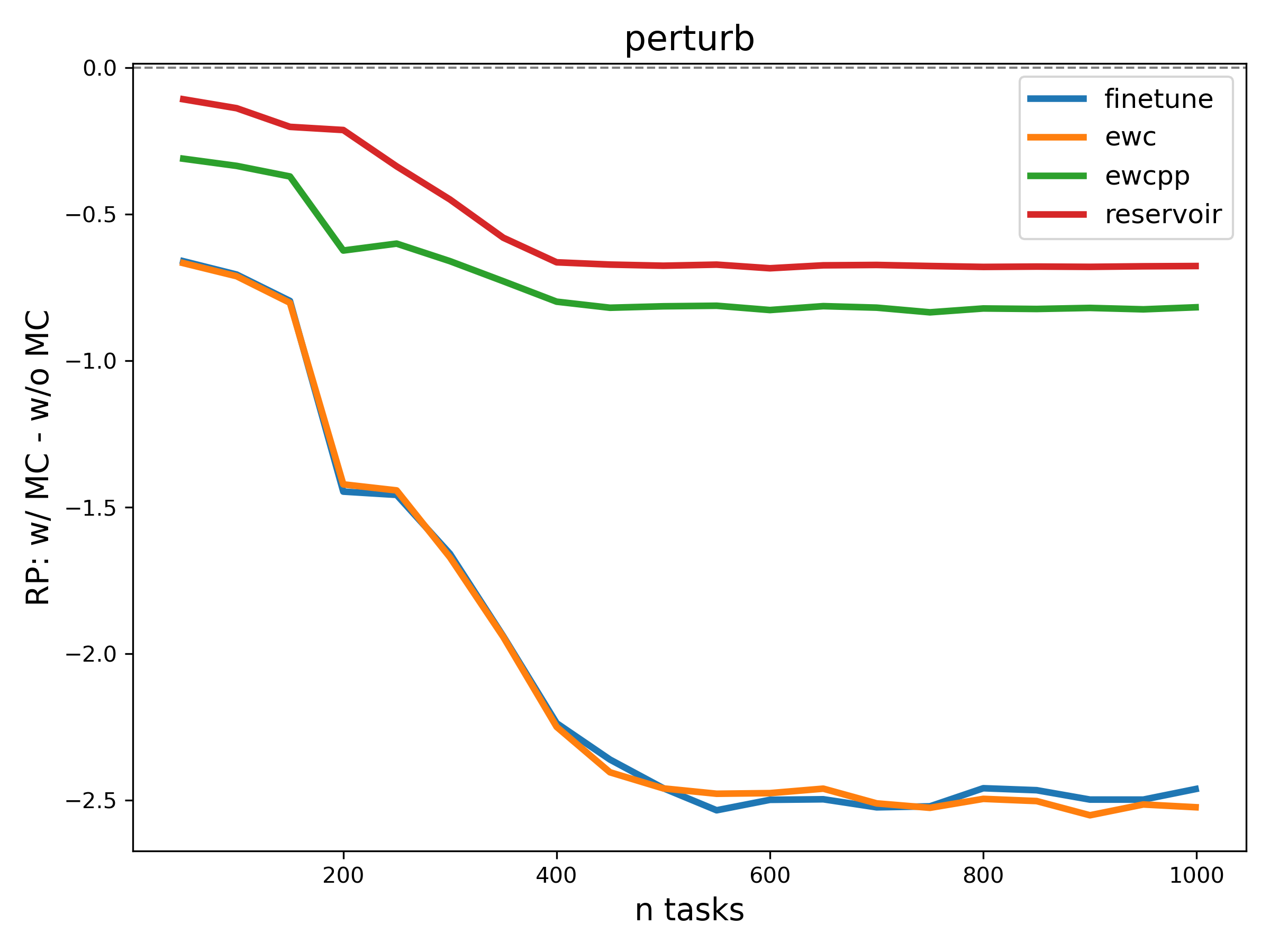}
        \caption{Synthetic task \texttt{perturb}.}
    \end{subfigure}
    \begin{subfigure}{.33\textwidth}
        \centering
        \includegraphics[width=\linewidth]{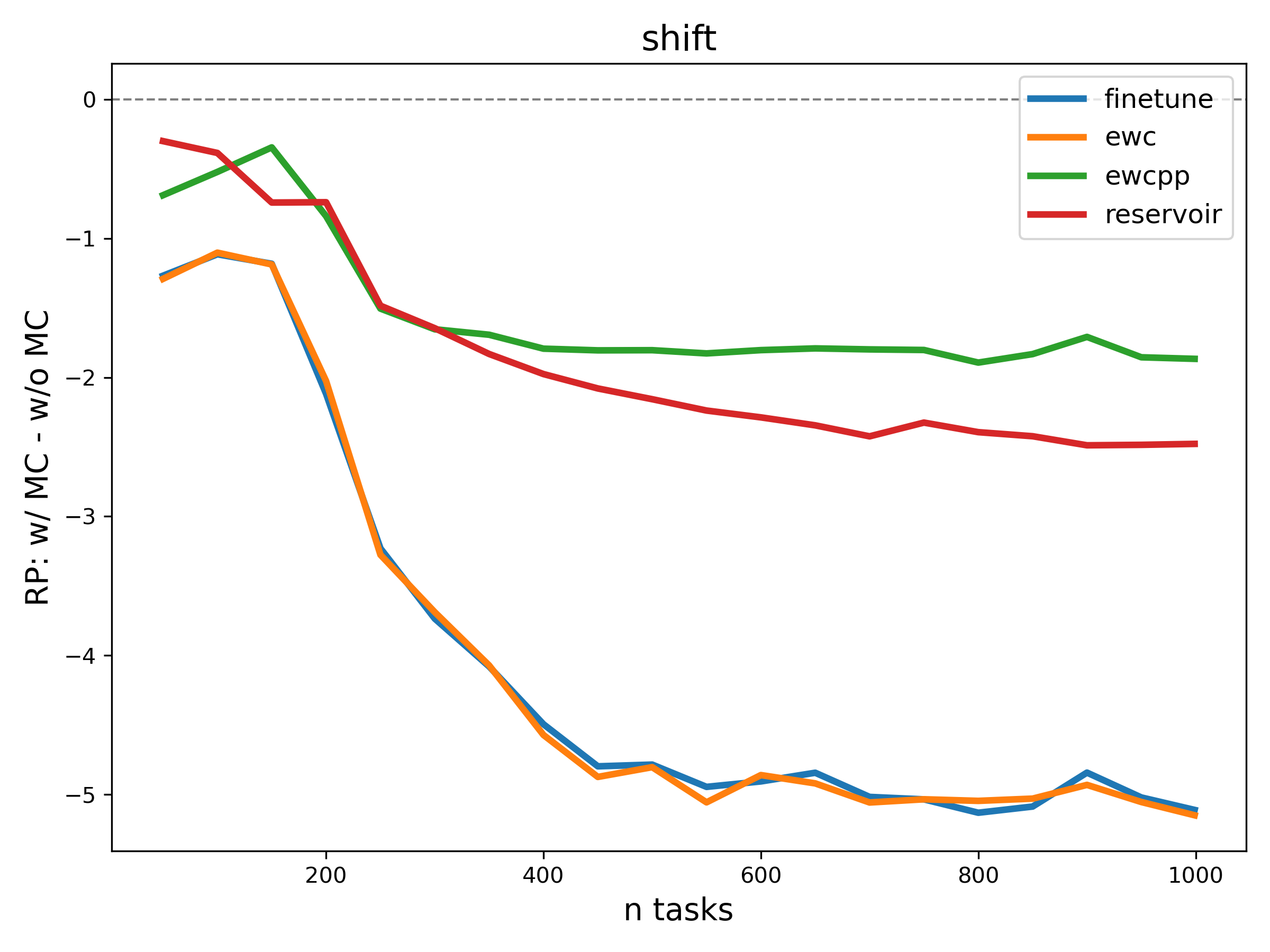}
        \caption{Synthetic task \texttt{shift}.}
    \end{subfigure}
    \caption{Difference between Continual Adam and Adam on synthetic linear regression tasks by number of tasks. X axis is number of tasks; Y axis is the difference between Continual Adam and Adam on RP metric (lower than zero means Continual Adam is better, since the metric for linear regression is lower better). The model is evaluated at $50, 100, 150, 200, ..., 1000$-th tasks.}
    \label{fig:syn}
\end{figure}

We show the difference between Continual AdamW and AdamW by number of seen tasks on \textbf{WSD-CL}, \textbf{VQA-CL}, and synthetic tasks in Figure \ref{fig:real} and \ref{fig:syn}. We see that the performance of Continual AdamW/Adam is mixed when number of seen tasks are small, while the gap between Continual AdamW/Adam and AdamW/Adam gets larger as the number of tasks grows, which indicates Continual AdamW/Adam is more suitable for continual learning scenarios with a large number of tasks.

\section{Implementation details}
All experiments are run on one Nvidia RTX3090 GPU. 
\subsection{WSD-CL}
\textbf{WSD-CL} dataset has 20399 tasks in total, 8566 of which are zero-shot tasks. The largest task is to disambiguate "be", which has 14204 target sentences for training, and on average each task has 9.28 target sentences to train. The word with most candidate senses to choose from is "break", which has 75 senses, and on average each word has 2.84 senses to choose from.

We use BERT-base checkpoint from huggingface library. We use batch size $12$, learning rate $1e-6$, and linear decay learning rate scheduler with warmup. For replay-based methods, we allocate a buffer of size $20000$, which is about 1\% of all data in \textbf{WSD-CL} dataset.

For Reservoir method, the weight for current task loss and replay loss are both $1$. Following notations in the original paper, for DER++ \citep{derpp}, we use $\alpha=1$, $\beta=1$; for EWC \citep{ewc}, we use $\lambda=2$; for EWC++ \citep{ewcpp}, we use $\alpha=0.9$, $\lambda=0.1$.

\subsection{VQA-CL}
\textbf{VQA-CL} dataset has 1000 tasks in total. The largest task is questions containing the word "appliance", which has 2396 questions, and on average each task has 137.86 question-image pairs. For all questions, the model chooses one answer from $1853$ candidate answers.

We modify the official implementation of Oscar model \footnote{https://github.com/microsoft/Oscar} for \textbf{VQA-CL} experiments. We use batch size $32$, learning rate $5e-5$, and linear decay learning rate scheduler with warmup. For replay-based methods, we allocate a buffer of size $1400$, which is about 1\% of all data in \textbf{VQA-CL} dataset.

For Reservoir method, the weight for current task loss and replay loss are both $1$. Following notations in the original paper, for DER++ \citep{derpp}, we use $\alpha=1$, $\beta=1$;  for EWC \citep{ewc}, we use $\lambda=2$; for EWC++ \citep{ewcpp}, we use $\alpha=0.9$, $\lambda=0.1$.

\subsection{Sequential-MNIST and sequential-CIFAR10}
We modify the mammoth implementation \footnote{https://github.com/aimagelab/mammoth} for sequential-MNIST and sequential-CIFAR10 experiments. We use batch size $10$, and search learning rate in $\{1e-5, 1e-4, 1e-3, 1e-2, 1e-1\}$ and report the best performance.

For Reservoir method, the weight for current task loss and replay loss are both $1$. Following notations in the original paper, for DER++ \citep{derpp}, we use $\alpha=1$, $\beta=1$;  for EWC \citep{ewc}, we use $\lambda=200$; for EWC++ \citep{ewcpp}, we use $\alpha=0.9$, $\lambda=1000$.

\subsection{5-task NLP continual learning}
We modify the official implementation of LAMOL \footnote{https://github.com/chho33/LAMOL} for the NLP tasks continual learning experiments. We use batch size $32$, learning rate $6.25e-5$, and linear decay learning rate scheduler with warmup.

For Reservoir method, the weight for current task loss and replay loss are both $1$. Following notations in the original paper, for DER++ \citep{derpp}, we use $\alpha=1$, $\beta=1$;  for EWC \citep{ewc}, we use $\lambda=2$; for EWC++ \citep{ewcpp}, we use $\alpha=0.9$, $\lambda=1$.

\section{Error bars}

We report error bars of results of \textbf{WSD-CL} and \textbf{VQA-CL} in the main paper in Table \ref{wsdcl_eb}, \ref{vqacl_eb}. Error bars are calculated from three runs with different random seeds.

\begin{table}[htbp]
\caption{Results on \textbf{WSD-CL}. "w/MC" columns are results of methods extended with our proposed algorithm, "w/o MC" columns are results of methods that reset optimizer states for each task. Error bars calculated from three independent runs using different random seeds.}
\label{wsdcl_eb}
\begin{tabular}{ccccccccc}
\hline
\multicolumn{1}{l}{} & \multicolumn{2}{c}{RP$\uparrow$}                                                                                        & \multicolumn{2}{c}{LP$\uparrow$}                                                                                        & \multicolumn{2}{c}{FGT$\downarrow$}                                                                                    & \multicolumn{2}{c}{BWT$\uparrow$}                                                                                        \\
\multicolumn{1}{l}{} & w/ MC                                                      & w/o MC                                                     & \multicolumn{1}{l}{w/ MC}                                  & \multicolumn{1}{l}{w/o MC}                                 & \multicolumn{1}{l}{w/ MC}                                 & \multicolumn{1}{l}{w/o MC}                                 & \multicolumn{1}{l}{w/ MC}                                  & \multicolumn{1}{l}{w/o MC}                                  \\ \hline
\vspace{0.2cm}
Finetune             & \begin{tabular}[c]{@{}c@{}}78.76\\ (\pm 0.10)\end{tabular} & \begin{tabular}[c]{@{}c@{}}61.63\\ (\pm 0.24)\end{tabular} & \begin{tabular}[c]{@{}c@{}}81.27\\ (\pm 0.01)\end{tabular} & \begin{tabular}[c]{@{}c@{}}79.53\\ (\pm 0.11)\end{tabular} & \begin{tabular}[c]{@{}c@{}}5.50\\ (\pm 0.09)\end{tabular} & \begin{tabular}[c]{@{}c@{}}25.98\\ (\pm 0.54)\end{tabular} & \begin{tabular}[c]{@{}c@{}}-2.51\\ (\pm 0.11)\end{tabular} & \begin{tabular}[c]{@{}c@{}}-17.90\\ (\pm 0.35)\end{tabular} \\
\vspace{0.2cm}
EWC                  & \begin{tabular}[c]{@{}c@{}}78.32\\ (\pm 0.63)\end{tabular} & \begin{tabular}[c]{@{}c@{}}72.48\\ (\pm 0.02)\end{tabular} & \begin{tabular}[c]{@{}c@{}}81.11\\ (\pm 0.09)\end{tabular} & \begin{tabular}[c]{@{}c@{}}79.56\\ (\pm 0.01)\end{tabular} & \begin{tabular}[c]{@{}c@{}}6.00\\ (\pm 0.56)\end{tabular} & \begin{tabular}[c]{@{}c@{}}10.65\\ (\pm 0.11)\end{tabular} & \begin{tabular}[c]{@{}c@{}}-2.78\\ (\pm 0.54)\end{tabular} & \begin{tabular}[c]{@{}c@{}}-7.08\\ (\pm 0.00)\end{tabular}  \\
\vspace{0.2cm}
EWC++                & \begin{tabular}[c]{@{}c@{}}78.10\\ (\pm 0.02)\end{tabular} & \begin{tabular}[c]{@{}c@{}}70.99\\ (\pm 4.20)\end{tabular} & \begin{tabular}[c]{@{}c@{}}80.61\\ (\pm 0.15)\end{tabular} & \begin{tabular}[c]{@{}c@{}}75.48\\ (\pm 0.03)\end{tabular} & \begin{tabular}[c]{@{}c@{}}5.47\\ (\pm 0.25)\end{tabular} & \begin{tabular}[c]{@{}c@{}}8.88\\ (\pm 5.66)\end{tabular}  & \begin{tabular}[c]{@{}c@{}}-2.52\\ (\pm 0.17)\end{tabular} & \begin{tabular}[c]{@{}c@{}}-4.49\\ (\pm 4.17)\end{tabular}  \\
\vspace{0.2cm}
Reservoir            & \begin{tabular}[c]{@{}c@{}}79.78\\ (\pm 0.08)\end{tabular} & \begin{tabular}[c]{@{}c@{}}74.48\\ (\pm 0.66)\end{tabular} & \begin{tabular}[c]{@{}c@{}}80.82\\ (\pm 0.01)\end{tabular} & \begin{tabular}[c]{@{}c@{}}78.84\\ (\pm 0.23)\end{tabular} & \begin{tabular}[c]{@{}c@{}}3.91\\ (\pm 0.11)\end{tabular} & \begin{tabular}[c]{@{}c@{}}9.15\\ (\pm 0.50)\end{tabular}  & \begin{tabular}[c]{@{}c@{}}-1.04\\ (\pm 0.07)\end{tabular} & \begin{tabular}[c]{@{}c@{}}-4.36\\ (\pm 0.42)\end{tabular}  \\
\vspace{0.2cm}
DER++                & \begin{tabular}[c]{@{}c@{}}79.12\\ (\pm 0.40)\end{tabular} & \begin{tabular}[c]{@{}c@{}}78.32\\ (\pm 0.05)\end{tabular} & \begin{tabular}[c]{@{}c@{}}79.48\\ (\pm 0.41)\end{tabular} & \begin{tabular}[c]{@{}c@{}}79.41\\ (\pm 0.01)\end{tabular} & \begin{tabular}[c]{@{}c@{}}4.02\\ (\pm 0.20)\end{tabular} & \begin{tabular}[c]{@{}c@{}}4.82\\ (\pm 0.13)\end{tabular}  & \begin{tabular}[c]{@{}c@{}}-0.36\\ (\pm 0.01)\end{tabular} & \begin{tabular}[c]{@{}c@{}}-1.10\\ (\pm 0.04)\end{tabular}  \\
\vspace{0.2cm}
AGEM                 & \begin{tabular}[c]{@{}c@{}}78.95\\ (\pm 0.36)\end{tabular} & \begin{tabular}[c]{@{}c@{}}69.37\\ (\pm 0.35)\end{tabular} & \begin{tabular}[c]{@{}c@{}}81.22\\ (\pm 0.01)\end{tabular} & \begin{tabular}[c]{@{}c@{}}79.56\\ (\pm 0.00)\end{tabular} & \begin{tabular}[c]{@{}c@{}}5.17\\ (\pm 0.37)\end{tabular} & \begin{tabular}[c]{@{}c@{}}16.50\\ (\pm 0.26)\end{tabular} & \begin{tabular}[c]{@{}c@{}}-2.27\\ (\pm 0.35)\end{tabular} & \begin{tabular}[c]{@{}c@{}}-10.19\\ (\pm 0.35)\end{tabular} \\ \hline
MTL                  & \multicolumn{2}{c}{\begin{tabular}[c]{@{}c@{}}83.72\\ (\pm 0.13)\end{tabular}}                                          & -                                                          & -                                                          & -                                                         & -                                                          & -                                                          & -                                                           \\ \hline
\end{tabular}
\end{table}

\begin{table}[htbp]
\caption{Results on \textbf{VQA-CL}. "w/MC" columns are results of methods extended with our proposed algorithm, "w/o MC" columns are results of methods that reset optimizer states for each task. Error bars calculated from three independent runs using different random seeds.}
\label{vqacl_eb}
\begin{tabular}{ccccccccc}
\hline
\multicolumn{1}{l}{} & \multicolumn{2}{c}{RP$\uparrow$}                                                                                        & \multicolumn{2}{c}{LP$\uparrow$}                                                                                        & \multicolumn{2}{c}{FGT$\downarrow$}                                                                                   & \multicolumn{2}{c}{BWT$\uparrow$}                                                                                      \\
\multicolumn{1}{l}{} & \multicolumn{1}{l}{w/ MC}                                  & \multicolumn{1}{l}{w/o MC}                                 & \multicolumn{1}{l}{w/ MC}                                  & \multicolumn{1}{l}{w/o MC}                                 & \multicolumn{1}{l}{w/ MC}                                 & \multicolumn{1}{l}{w/o MC}                                & \multicolumn{1}{l}{w/ MC}                                 & \multicolumn{1}{l}{w/o MC}                                 \\ \hline
\vspace{0.2cm}
Finetune             & \begin{tabular}[c]{@{}c@{}}69.98\\ (\pm 0.26)\end{tabular} & \begin{tabular}[c]{@{}c@{}}64.99\\ (\pm 0.15)\end{tabular} & \begin{tabular}[c]{@{}c@{}}67.29\\ (\pm 0.25)\end{tabular} & \begin{tabular}[c]{@{}c@{}}65.69\\ (\pm 0.09)\end{tabular} & \begin{tabular}[c]{@{}c@{}}6.17\\ (\pm 0.04)\end{tabular} & \begin{tabular}[c]{@{}c@{}}8.59\\ (\pm 0.08)\end{tabular} & \begin{tabular}[c]{@{}c@{}}2.68\\ (\pm 0.01)\end{tabular} & \begin{tabular}[c]{@{}c@{}}-0.70\\ (\pm 0.06)\end{tabular} \\
\vspace{0.2cm}
EWC                  & \begin{tabular}[c]{@{}c@{}}69.62\\ (\pm 0.19)\end{tabular} & \begin{tabular}[c]{@{}c@{}}67.27\\ (\pm 0.10)\end{tabular} & \begin{tabular}[c]{@{}c@{}}66.16\\ (\pm 0.24)\end{tabular} & \begin{tabular}[c]{@{}c@{}}66.16\\ (\pm 0.02)\end{tabular} & \begin{tabular}[c]{@{}c@{}}5.85\\ (\pm 0.04)\end{tabular} & \begin{tabular}[c]{@{}c@{}}7.84\\ (\pm 0.03)\end{tabular} & \begin{tabular}[c]{@{}c@{}}3.46\\ (\pm 0.03)\end{tabular} & \begin{tabular}[c]{@{}c@{}}1.11\\ (\pm 0.05)\end{tabular}  \\
\vspace{0.2cm}
EWC++                & \begin{tabular}[c]{@{}c@{}}69.40\\ (\pm 0.58)\end{tabular} & \begin{tabular}[c]{@{}c@{}}64.07\\ (\pm 0.08)\end{tabular} & \begin{tabular}[c]{@{}c@{}}66.79\\ (\pm 0.17)\end{tabular} & \begin{tabular}[c]{@{}c@{}}65.29\\ (\pm 0.02)\end{tabular} & \begin{tabular}[c]{@{}c@{}}6.49\\ (\pm 0.37)\end{tabular} & \begin{tabular}[c]{@{}c@{}}9.15\\ (\pm 0.07)\end{tabular} & \begin{tabular}[c]{@{}c@{}}2.61\\ (\pm 0.41)\end{tabular} & \begin{tabular}[c]{@{}c@{}}-1.22\\ (\pm 0.10)\end{tabular} \\
\vspace{0.2cm}
Reservoir            & \begin{tabular}[c]{@{}c@{}}71.61\\ (\pm 0.16)\end{tabular} & \begin{tabular}[c]{@{}c@{}}67.44\\ (\pm 0.14)\end{tabular} & \begin{tabular}[c]{@{}c@{}}68.51\\ (\pm 0.30)\end{tabular} & \begin{tabular}[c]{@{}c@{}}67.05\\ (\pm 0.10)\end{tabular} & \begin{tabular}[c]{@{}c@{}}5.51\\ (\pm 0.05)\end{tabular} & \begin{tabular}[c]{@{}c@{}}7.45\\ (\pm 0.11)\end{tabular} & \begin{tabular}[c]{@{}c@{}}3.09\\ (\pm 0.14)\end{tabular} & \begin{tabular}[c]{@{}c@{}}0.39\\ (\pm 0.24)\end{tabular}  \\
\vspace{0.2cm}
DER++                & \begin{tabular}[c]{@{}c@{}}71.37\\ (\pm 0.19)\end{tabular} & \begin{tabular}[c]{@{}c@{}}66.63\\ (\pm 0.06)\end{tabular} & \begin{tabular}[c]{@{}c@{}}68.14\\ (\pm 0.38)\end{tabular} & \begin{tabular}[c]{@{}c@{}}66.61\\ (\pm 0.08)\end{tabular} & \begin{tabular}[c]{@{}c@{}}5.47\\ (\pm 0.05)\end{tabular} & \begin{tabular}[c]{@{}c@{}}7.18\\ (\pm 0.01)\end{tabular} & \begin{tabular}[c]{@{}c@{}}3.22\\ (\pm 0.19)\end{tabular} & \begin{tabular}[c]{@{}c@{}}0.02\\ (\pm 0.02)\end{tabular}  \\
\vspace{0.2cm}
AGEM                 & \begin{tabular}[c]{@{}c@{}}69.98\\ (\pm 0.14)\end{tabular} & \begin{tabular}[c]{@{}c@{}}64.57\\ (\pm 0.20)\end{tabular} & \begin{tabular}[c]{@{}c@{}}66.69\\ (\pm 0.25)\end{tabular} & \begin{tabular}[c]{@{}c@{}}65.52\\ (\pm 0.18)\end{tabular} & \begin{tabular}[c]{@{}c@{}}6.19\\ (\pm 0.04)\end{tabular} & \begin{tabular}[c]{@{}c@{}}8.93\\ (\pm 0.13)\end{tabular} & \begin{tabular}[c]{@{}c@{}}3.30\\ (\pm 0.11)\end{tabular} & \begin{tabular}[c]{@{}c@{}}-0.95\\ (\pm 0.03)\end{tabular} \\ \hline
MTL                  & \multicolumn{2}{c}{\begin{tabular}[c]{@{}c@{}}73.98\\ (\pm 0.15)\end{tabular}}                                          & -                                                          & -                                                          & -                                                         & -                                                         & -                                                         & -                                                          \\ \hline
\end{tabular}
\end{table}
\newpage
\bibliography{bibliography}
\bibliographystyle{plainnat}